# Does BERT look at sentiment lexicon?

Elena Razova[1[0000-0001-5557-5432]], Sergey Vychegzhanin[1[0000-0001-6456-7856]],
and Evgeny Kotelnikov[1,2[0000-0001-9745-1489]]

[1] Vyatka State University, Kirov, Russia
[2] ITMO University, Saint-Petersburg, Russia
`{razova.ev, vychegzhaninsv, kotelnikov.ev}@gmail.com`

**Abstract.** The main approaches to sentiment analysis are rule-based methods and machine learning, in particular, deep neural network models with the Transformer architecture, including BERT. The performance of neural network models in the tasks of sentiment analysis is superior to the performance of rule-based methods. The reasons for this situation remain unclear due to the poor interpretability of deep neural network models. One of the main keys to understanding the fundamental differences between the two approaches is the analysis of how sentiment lexicon is taken into account in neural network models. To this end, we study the attention weights matrices of the Russian-language RuBERT model. We fine-tune RuBERT on sentiment text corpora and compare the distributions of attention weights for sentiment and neutral lexicons. It turns out that, on average, 3/4 of the heads of various model variants statistically pay more attention to the sentiment lexicon compared to the neutral one.

## 1 Introduction

There are two main approaches to sentiment analysis of texts [3]: lexicon-based (or rule-based) and machine learning, in particular, using deep neural network models based on the Transformer architecture [31], including BERT [9]. The lexicon-based methods are high speed, easy to implement, they require no training data and a long learning process, and their results are easy to interpret [29]. However, neural network models achieve higher classification performance, for example, all state-of-the-art results for well-known sentiment analysis tasks (for example, SST, Yelp reviews, IMDB reviews, Amazon reviews) are achieved by neural network models[1].

Despite much work on the study of deep neural network mechanisms [2, 32], the reasons for this success still remain incomprehensible. This is due to the high complexity of interpreting the neural network models containing hundreds of millions of parameters.

---

[1] https://paperswithcode.com/task/sentiment-analysis.

Taking the sentiment lexicon into consideration is one of the key aspects for understanding the differences between the lexicon-based approach and deep learning in sentiment analysis tasks. This paper explores the extent to which the BERT neural network model pays attention to sentiment words. For this purpose, we construct the distribution of attention weights for the Russian-language neural network model RuBERT [18] for various subsets of words – a sentiment lexicon in general, positive and negative lexicons, as well as neutral one. Distributions are constructed for three variants of RuBERT models trained on three sentiment corpora (ROMIP 2012 News, SentiRuEval-2015 Banks and RuSentiment). We calculate and analyze distances between the constructed distributions based on Kullback-Leibler divergence. The Wilcoxon signed-rank test is used to test the significance of the distance discrepancy. The analysis made it possible to conclude that, on average, 3/4 of the heads of various variants of the RuBERT model statistically pay more attention to the sentiment lexicon compared to the neutral one.

The contribution of this article is as follows:

— we propose a method that analyzes how neural network models take into account various lexical subsets based on the distribution of attention weights and calculating the Kullback-Leibler distance between them;
— we analyze the accounting of sentiment and neutral words in the RuBERT model in the sentiment analysis task for three sentiment corpora;
— we conclude that on average RuBERT heads pay attention to sentiment words to a greater extent than to neutral ones, and this difference is statistically significant.

## 2 Previous Work

Interpretation of neural network models is carried out using three main approaches [2]: structural analysis, behavioral analysis and visualization.

Structural analysis examines various components of the neural network, such as word embeddings, sentence embeddings, attention weights, hidden layers, etc. [5, 37, 36]. The result of the analysis is to determine the role of the components in the neural network and to specify the type of information that these components can take into consideration.

Behavioral analysis consists in studying a neural network model on a set of test corpora, each of them reflects specific linguistic phenomena [1, 19]. Usually, most reference datasets are taken from text collections reflecting the natural frequency distribution of linguistic phenomena. Such sets are useful for assessing the average accuracy of a model, but may not reflect a wide range of linguistic phenomena. An alternative approach to assessing performance is to use different sets of tests, which can be created to investigate models for different tasks, natural languages, corpus sizes, etc.

Visualization is a complementary approach preceding structural and behavioral analysis [5, 8, 10]. Visualization helps to generate hypotheses about the behavior of a model or a dataset and to understand complex concepts.

Our research is carried out within the framework of structural analysis and the closest works are [5, 36].

Cao et al. [5] developed a differentiable masking method that allowed us to find out what different layers of the model "know" about the input data and where the prediction information is stored in different layers. The BERT$_{BASE}$ model was investigated for sentiment classification on the Stanford Sentiment Treebank (SST) [27]. Most layers have been found to rely on very positive or very negative words. In contrast to our work, the authors consider layers as a whole, without going down to the level of individual attention heads and specific weight distributions. They use annotated words in the SST as sentiment words, rather than a universal sentiment dictionary, as in our study.

Wu et al. [36] investigated the self-attention mechanism in the Transformer using the Layer-wise Attention Tracing method. SST and Stanford Emotional Narratives Dataset were used as corpora [24]. The authors have shown that attention weights have the highest values for very positive and very negative words. For this purpose, the dictionary by Warriner et al. was used, containing about 14,000 words with emotional valence [33]. The paper has also analyzed the proportion of attention given by individual heads to the sentiment words. In contrast to our work, [36] does not use BERT, but other Transformer-based models. They also use a sentiment dictionary, but to get the weight of the token, they sum up all the attention weights without considering the distribution of the weights. In our work, we build such distributions, which allows us to compare the weights for sentiment and neutral lexicons in more detail.

## 3    Interpretation Method

Our work aims to investigate the extent to which BERT-type models [9] pay attention to sentiment words. The research is based on the following hypothesis: BERT-type models pay different attention to sentiment and neutral lexicons. To test this hypothesis, the significance of the difference between the distributions of attention weights of sentiment (positive and negative) words and neutral words has been evaluated. The distances between the distributions are calculated based on the Kullback-Leibler divergence.

The interpretation method involves four steps.

**Step 1.** Building an average attention matrix.

BERT uses WordPiece tokenization [35], whereby part of words is split into several tokens (subwords)[2]. A word-level analysis of attention heads is required to calculate the attention that BERT assigns to sentiment words. Therefore, we convert the attention matrices of individual heads "token-token" into attention matrices "word-word". For this purpose, we implement the following procedure (see Fig. 1).

1. We summarize the attention weights directed to a tokenized word, by its tokens.
2. We take the average value of the attention weights over its tokens, directed from the split word to other tokens. This transformation preserves the property of the attention matrix, which means that the sum of the attention weights for each word is equal to one.

---

[2] RuBERT model for Russian uses BPE (Byte Pair Encoding) tokenization [26].

3. For each attention head and word, we find the average attention weight given to the word in the text.
4. We combine the separated words and lemmatize[3] all words of the text.

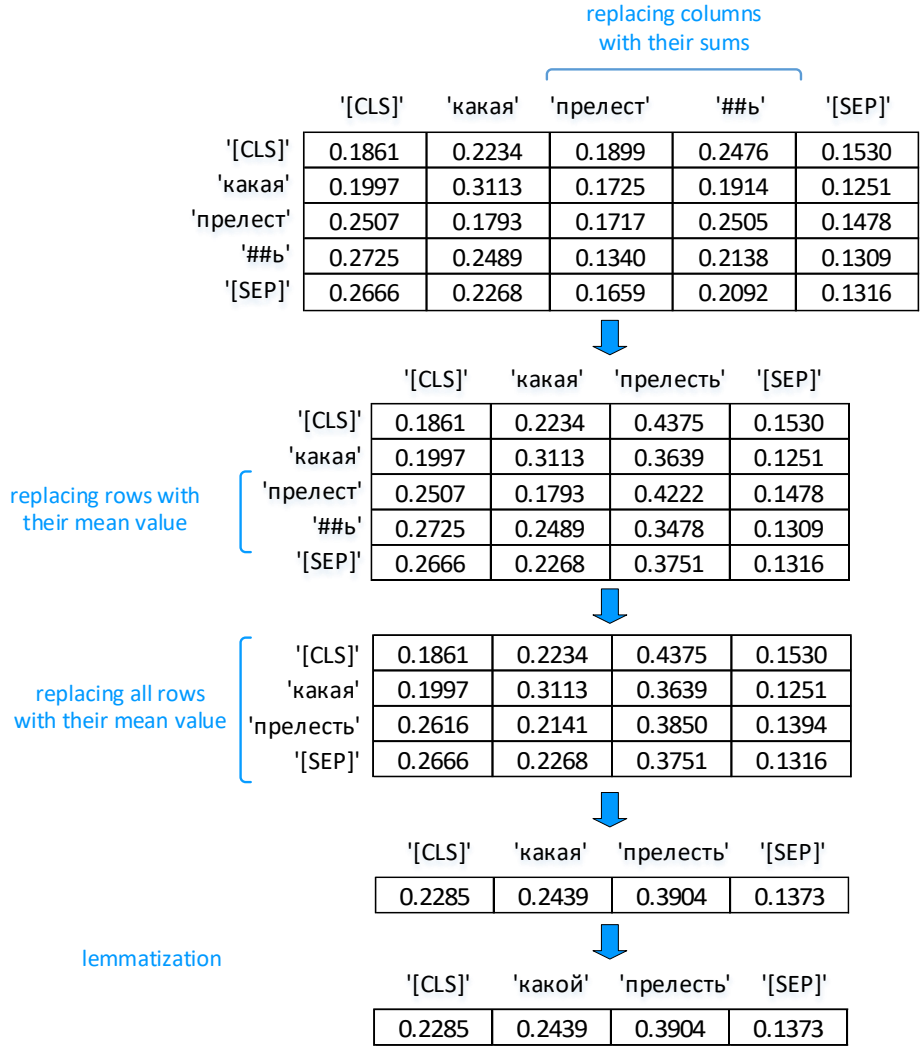

**Fig. 1.** Building an average attention matrix.

Thus, the first step results in finding the average attention weight given to this word in the text for each attention head and each word.

**Step 2.** Calculation of the distribution of attention weights for sentiment ($W_s$) and non-sentiment (neutral) words ($W_{ns}$) of the corpus. By non-sentiment words, we mean words that are not included in the sentiment dictionary. Sentiment words of the corpus are defined using the sentiment dictionary (see Subsection 4.2).

---

[3] The pymorphy2 [12] library was used for lemmatization.

To confirm the hypothesis that BERT pays different attention to sentiment and neutral lexicons, it is necessary to establish the difference in the distribution of attention weights of these types of lexicon. However, this difference can be accidental. To reduce the likelihood of a false conclusion, we ran 10 tests: on each test, we generated a random subset of neutral words for each attention head. The size of each random subset coincided with the number of sentiment words in the corpus.

We introduce the following notation:

— $W_p$ – set of positive corpus words;
— $W_n$ – set of negative corpus words;
— $W_s = W_p \cup W_n$ – set of sentiment corpus words;
— $W_{ns}$ – set of non-sentiment (neutral) corpus words;
— $W_{e,i}$ – $i$-th random subset of neutral words of the set $W_{ns}$: $W_{e,i} \subset W_{ns}$, $|W_{e,i}| = |W_s|$, $i = [1,10]$;
— $W_{o,i} = W_{ns} / W_{e,i}$ – set of neutral words after eliminating the $i$-th random subset.

At each of the ten tests, we determined the distribution of attention weights for words of five sets: sentiment ($W_p$, $W_n$, $W_s$) and neutral ($W_{e,i}$, $W_{o,i}$) words. The distributions of weights for all sets were calculated over one hundred value intervals (from 0 to 1 with a step of 0.01).

At the next step, the distances between the distributions of attention weights for sentiment ($W_p$, $W_n$, $W_s$) and neutral ($W_{o,i}$) words will be calculated, as well as the distances between the distributions of attention weights for subsets of neutral words $W_{e,i}$ and $W_{o,i}$. Comparing such distances enables to make a statistically significant conclusion about the difference in the distribution of attention between sentiment and neutral lexicons.

**Step 3.** Calculation of the Kullback-Leibler divergence between the distributions of attention weights of different word sets.

The distance between the distributions of attention weights was calculated based on the Kullback-Leibler divergence. The Kullback-Leibler divergence of the $Q$ distribution relative to the $P$ distribution (or "distance" from $P$ to $Q$) is defined as follows [22, p. 34]:

$$D(P||Q) = \sum_{i=1}^{n} p_i log \frac{p_i}{q_i}. \tag{1}$$

For each set of neutral words $W_{o,i}$ and each $h$-th attention head, the following Kullback-Leibler distances are found:

$$D_{o,s,i}^h = D(P_{W_{o,i}}^h || P_{W_s}^h),$$
$$D_{o,p,i}^h = D(P_{W_{o,i}}^h || P_{W_p}^h), \tag{2}$$

$$D_{o,n,i}^h = D(P_{W_{o,i}}^h || P_{W_n}^h),$$

$$D_{o,e,i}^h = D(P_{W_{o,i}}^h || P_{W_e}^h),$$

where $P_{W_{o,i}}^h$, $P_{W_s}^h$, $P_{W_p}^h$, $P_{W_n}^h$, $P_{W_e}^h$ – distribution of attention weights, respectively, of the sets of words $W_{o,i}$, $W_s$, $W_p$, $W_n$, $W_e$ for the head $h$.

Next, for each attention head, the average Kullback-Leibler distances are found:

$$D_{o,s}^h = \frac{\sum_{i=1}^{10} D_{o,s,i}^h}{10},$$

$$D_{o,p}^h = \frac{\sum_{i=1}^{10} D_{o,p,i}^h}{10},$$

$$D_{o,n}^h = \frac{\sum_{i=1}^{10} D_{o,n,i}^h}{10},$$

$$D_{o,e}^h = \frac{\sum_{i=1}^{10} D_{o,e,i}^h}{10}.$$

(3)

**Step 4.** Testing the significance of differences between Kullback-Leibler distances for sentiment and neutral words.

To test the significance of the differences, we formulated two hypotheses:

— $H_0$ – the discrepancies between the Kullback-Leibler distances of sentiment and neutral words are random.
— $H_1$ – the discrepancies between the Kullback-Leibler distances of sentiment and neutral words are non-random.

To test these hypotheses, the Wilcoxon signed-rank test was used, which is designed to test the differences between two samples of independent measurements [34].

The Wilcoxon signed-rank test was applied at a significance level of $p = 0.05$ to evaluate hypotheses for the following pairs of Kullback-Leibler distances obtained in the previous step: $\langle D_{o,s}^h, D_{o,e}^h \rangle$, $\langle D_{o,p}^h, D_{o,e}^h \rangle$, $\langle D_{o,n}^h, D_{o,e}^h \rangle$.

Wilcoxon's test makes it possible to draw a conclusion only about the statistical significance of differences between samples of distances, but not about which distances prevail on average. For this purpose, we calculated the expected value of sets of distances.

Thus, the proposed interpretation method allows us to test and statistically substantiate the hypothesis that the BERT model pays different attention to sentiment and neutral lexicons, as well as to determine which lexicon is given more attention to.

# 4 Resources and Models

## 4.1 Text Corpora

In our experiments, we used three Russian-language corpora, annotated by sentiment: ROMIP 2012 News, SentiRuEval-2015 Banks and RuSentiment. We had two criteria when choosing corpora for analysis: 1) corpora should differ in the type of texts; 2) lexicon-based methods should show significantly lower performance for these corpora than neural network models.

The news corpus was prepared for the ROMIP 2012 sentiment analysis competition [7]. The corpus includes fragments of direct and indirect speech from news. The corpus of tweets about banks was prepared for the SentiRuEval-2015 sentiment analysis competition [21]. RuSentiment corpus includes posts on VKontakte [25]. We only used training parts of the corpora. The characteristics of the corpora are presented in Table 1.

**Table 1.** Characteristics of corpora, annotated by sentiment.
The length of the texts is specified in tokens (mean±std)[4].

| Corpora | Total | Positive | Negative | Neutral | Length |
|---------|-------|----------|----------|---------|--------|
| ROMIP | 4,260 | 1,115 (26%) | 1,864 (44%) | 1,281 (30%) | 35±28 |
| SentiRuEval | 4,883 | 354 (7%) | 1,059 (22%) | 3,470 (71%) | 10±5 |
| RuSentiment | 24,124 | 9,170 (38%) | 3,654 (15%) | 11,300 (47%) | 13±17 |

Our experiments showed that for these corpora the difference between the results of the lexicon-based method (we used the version of the SO-CAL method [28] adapted for the Russian language) and the RuBERT model is the largest in comparison to other corpora. In particular, for ROMIP 2012 News, the difference in the macro F1-score for a three-class problem was 21 percentage points (p.p.), for SentiRuEval-2015 Banks – 20 p.p., for RuSentiment – 27 p.p. [16].

## 4.2 Sentiment Dictionary

The sentiment dictionary must be highly accurate and complete. To create such a dictionary, we combined 9 publicly available Russian sentiment dictionaries [14, 16]: RuSentiLex [20], Word Map [17], SentiRusColl [15], EmoLex [23], LinisCrowd [11], Blinov's lexicon [4], Kotelnikov's lexicon [13], Chen-Skiena's lexicon [6], Tutubalina's lexicon [30]. The characteristics of the lexicons are shown in Table 2.

---

4  Tokenization was carried out using NLTK (https://www.nltk.org).

**Table 2.** The characteristics of sentiment lexicons.

| Lexicon | Total | Positive elements | | Negative elements | |
|---|---|---|---|---|---|
| | | # | % | # | % |
| RuSentiLex | 12,560 | 3,258 | 25.9% | 9,302 | 74.1% |
| Word Map | 11,237 | 4,491 | 40.0% | 6,746 | 60.0% |
| SentiRusColl | 6,538 | 3,981 | 60.9% | 2,557 | 39.1% |
| EmoLex | 4,600 | 1,982 | 43.1% | 2,618 | 56.9% |
| LinisCrowd | 3,986 | 1,126 | 28.2% | 2,860 | 71.8% |
| Blinov's lexicon | 3,524 | 1,611 | 45.7% | 1,913 | 54.3% |
| Kotelnikov's lexicon | 3,206 | 1,028 | 32.1% | 2,178 | 67.9% |
| Chen-Skiena's lexicon | 2,604 | 1,139 | 43.7% | 1,465 | 56.3% |
| Tutubalina's lexicon | 2,442 | 1,032 | 42.3% | 1,410 | 57.7% |

The final dictionary included only those words that are included in at least $N$ source dictionaries. In accordance with our preliminary experiments, the optimal value for sentiment analysis based on the SO-CAL lexicon-based method is demonstrated by a dictionary with $N = 4$.

The final dictionary contains 2,313 words, including 823 positive (35.6%) and 1,490 negative (64.4%) words.

### 4.3 RuBERT Model

In our study the Russian-language neural network pre-trained model RuBERT [18] is used. The model was initialized on the basis of the multilingual version of BERT$_{BASE}$ and trained on the Russian part of Wikipedia and news articles.

The RuBERT model was fine-tuned (separately) on ROMIP 2012 News, SentiRuEval-2015 Banks and RuSentiment with the following parameters: number of epochs 5, batch size 8, learning rate $10^{-6}$.

## 5 Results and Discussion

In the experimental part of the study, we applied the interpretation method proposed in Section 3 for three versions of the RuBERT model trained on three corpora: ROMIP 2012 News, SentiRuEval-2015 Banks and RuSentiment. All attention heads were examined only on the last (12th) layer.

**Step 1.** At the first step, we built the average attention matrices for each RuBERT variant.

**Step 2.** We calculated the distributions of attention weights of the sentiment ($W_s$) and neutral words ($W_{ns}$) of the corpus over one hundred value intervals (from 0 to 1 with a step of 0.01). An example of the distribution of attention weights for the seventh head of the last layer for the ROMIP 2012 News corpus is shown in Fig. 2.

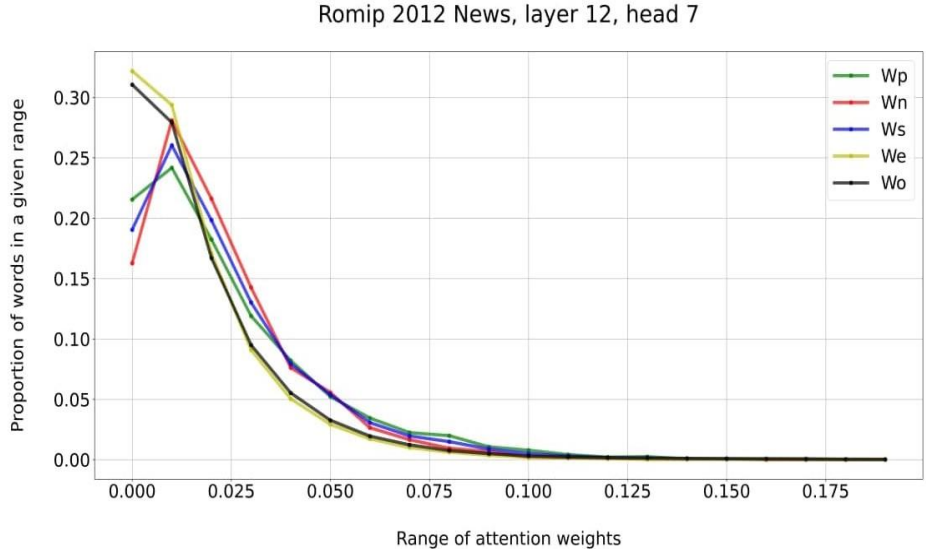

**Fig. 2.** Distribution of attention weights of sentiment and neutral words for the ROMIP 2012 News corpus for head #7 of the last layer.

**Step 3.** We calculated the Kullback-Leibler distances between the distributions of the attention weights of words.

**Step 4.** We tested the significance of the differences between Kullback-Leibler distances for sentiment and neutral words.

Table 3 shows the test results. Each cell contains two signs. The first sign is responsible for the result of checking the significance of the difference between the attention weights distributions of the corresponding sets of sentiment words ($sent - D_{o,s}^h$, $pos - D_{o,p}^h$, $neg - D_{o,n}^h$) and a set of neutral words $D_{o,e}^h$: "plus" indicates that the difference is significant, "minus" – not significant. The second sign is responsible for the ratio of the expected value of the corresponding distributions: "plus" means that the expected value of the attention weights distribution of the set of sentiment words is greater than the expected value of the distribution of the weights of the set of neutral words $D_{o,e}^h$; "minus" – less. At the bottom of the table, the number of "plus-plus" situations is given – the discrepancies are significant (non-random) and the expected value of the attention weights of the sentiment words is greater than the expected value of the attention weights of neutral words.

Table 3 shows that the difference of the attention weights distributions of sentiment words from the distributions of neutral words in 76.85% of cases is not random (the first sign in a pair is "plus"), and in 75% of cases, apart from the non-randomness of differences, the expected value of sentiment words is greater than the expected value of neutral words ("plus-plus" situation).

**Table 3.** Results of testing the significance of differences between Kullback-Leibler distances for sentiment and neutral words.

| head | ROMIP 2012 News | | | SentiRuEval-2015 Banks | | | RuSentiment | | |
|---|---|---|---|---|---|---|---|---|---|
| | sent | pos | neg | sent | pos | neg | sent | pos | neg |
| 0 | + + | + + | − + | + − | + − | − − | + + | + + | + + |
| 1 | + + | − + | + + | + + | − + | + + | + + | + + | − + |
| 2 | − + | + + | + + | − + | + + | + + | + + | + + | + + |
| 3 | + + | + + | + + | + + | + + | + + | + + | + + | − + |
| 4 | + + | + + | + + | + + | + + | + + | − + | + + | − + |
| 5 | + + | + + | + + | + + | + + | + + | + + | − + | + + |
| 6 | − + | + + | + + | − + | + + | + + | + + | + + | + + |
| 7 | + + | + + | + + | + + | + + | + + | + + | + + | + + |
| 8 | − + | + + | + + | − + | + + | + + | + + | + + | − + |
| 9 | + + | − + | + + | + + | − + | + + | + + | + + | + + |
| 10 | + + | + + | + + | + + | + + | + + | + + | + + | + + |
| 11 | + + | − + | − + | + + | − + | − + | + + | − + | − + |
| + + | 9 | 9 | 10 | 8 | 8 | 10 | 11 | 10 | 6 |

The extent of attention for positive and negative lexicons, as well as for sentiment lexicon in general, does not differ notably. Table 3 shows that the picture is approximately the same for all three corpora (only for RuSentiment fewer heads pay attention to negative words than on average).

Interestingly, head #7 for all the corpora highlights sentiment words, both in general and positive / negative individually.

We have also noticed an interesting effect that is observed for all the three corpora: the expected value of the attention weights to neutral words decreases monotonically from the first head to the last.

We also looked at some examples of how attention weights to sentiment words affect the final decision of the model (Fig. 3).

In the first example (the RuSentiment corpus) RuBERT pays maximum attention to the sentiment positive word "шикарный" and the classification result coincides with the true label. However, in the second example (also from the RuSentiment corpus) RuBERT also pays maximum attention to the positive word "прикольный", but nevertheless decides on the neutrality of the text (with true label = "positive").

Thus, despite the high attention weights in relation to sentiment words, RuBERT does not necessarily make the final decision based only on them. Obviously, the decision-making process is more complex and requires further study.

## 6 Conclusion

Thus, we can conclude that 3/4 of heads, on average, pay more attention to sentiment words than neutral ones. The obtained results are consistent with studies [5, 36], but it is for the first time that we have obtained quantitative estimates of the degree of attention to sentiment words, verified on the basis of the statistical criterion.

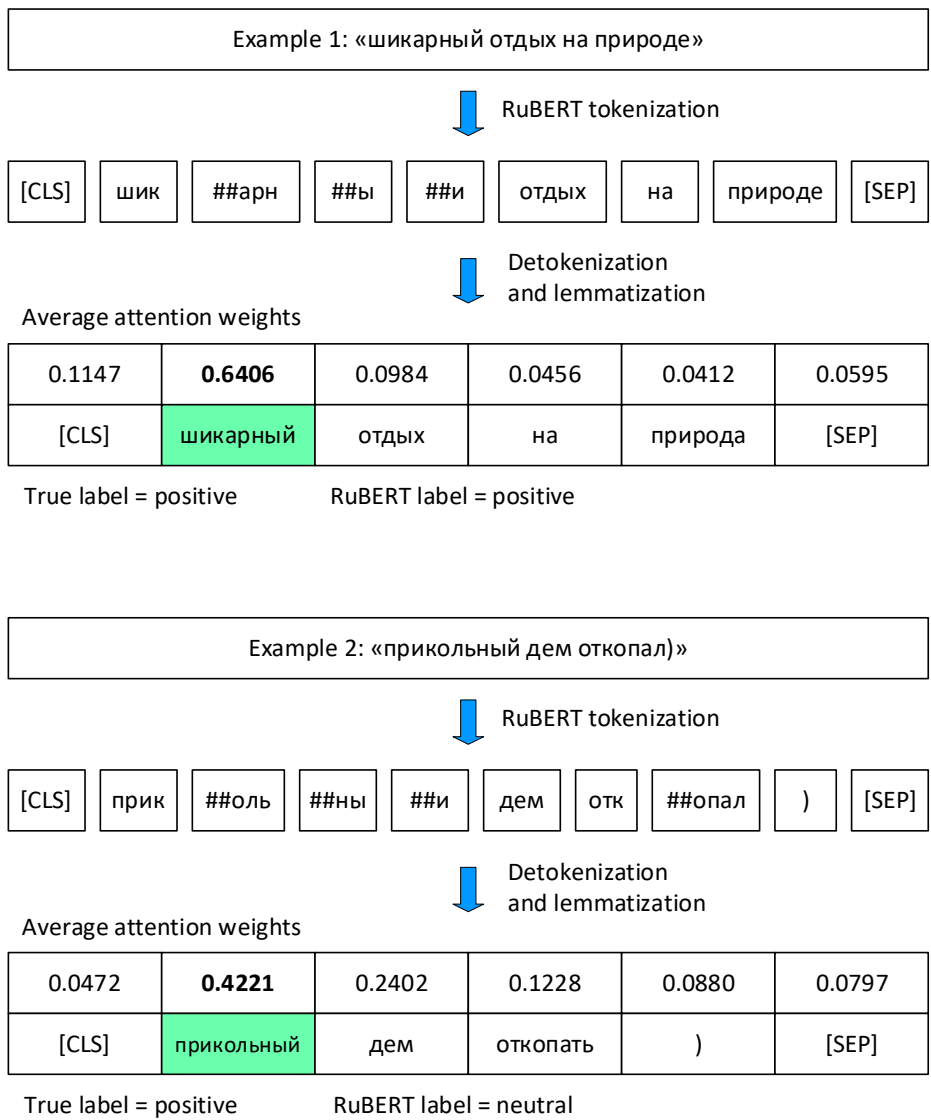

**Fig. 3.** Examples of texts and RuBERT attention weights.

In the future, we plan:

— to expand the sentiment dictionary – some sentiment words were not found due to insufficient size of the dictionary;
— to conduct a similar study on other corpora, annotated by sentiment;
— to improve the procedure for detokenization – some words after detokenization lost the letter "й" due to the peculiarities of the RuBERT tokenizer;

— to increase the number of tests from 10 to 100;
— to conduct an in-depth study of the attention weights, without averaging attention for words;
— to examine attention on different layers.